\documentclass{article}
\usepackage[utf8]{inputenc}
\usepackage{microtype}
\usepackage{graphicx}
\usepackage{booktabs} % for professional tables
\usepackage{comment}
\usepackage{amsmath}
\usepackage{amsfonts}
\usepackage{amssymb}
\usepackage{times}
\usepackage{calligra}
\usepackage{calrsfs}
\usepackage{paralist, tabularx}
\usepackage[numbers]{natbib}
\usepackage{hyperref}
\usepackage{authblk}
\usepackage{amsmath, amsfonts}
\usepackage{mathtools}
\usepackage{commath}
\usepackage{multirow}
\usepackage{color}
\usepackage{float}
\usepackage{subfig}
\usepackage{url}            % simple URL typesetting
\usepackage{wrapfig}
\usepackage{bbold}
\usepackage{algorithm}
\usepackage[noend]{algpseudocode}

\newcommand{\bx}{\mathbf{x}}

\def\bx{{\boldsymbol{x}}}

\def\bg{{\boldsymbol{g}}}
\def\bu{{\boldsymbol{u}}}
\def\btheta{{\boldsymbol{\theta}}}
\newcommand{\R}{\mathbb{R}}
\def\L{\mathcal{L}}

\newcommand{\argmax}{\operatornamewithlimits{argmax}}
\newcommand{\argmin}{\operatornamewithlimits{argmin}}

\newtheorem{theorem}{Theorem}

\title{Query-Efficient Hard-label Black-box Attack: \\An Optimization-based Approach}

\author{Minhao Cheng\textsuperscript{1},\enskip Thong Le\textsuperscript{1},\enskip Pin-Yu Chen\textsuperscript{3},\enskip Jinfeng Yi\textsuperscript{2},\enskip Huan Zhang\textsuperscript{1},\enskip Cho-Jui Hsieh\textsuperscript{1}\\
\textsuperscript{1}Department of Computer Science, University of California,  Davis, CA 95616 \\
\textsuperscript{2}JD AI Research, Beijing, China\\
\textsuperscript{3}IBM Research, Yorktown Heights, NY 10598 \\
\small{\texttt{mhcheng@ucdavis.edu, thmle@ucdavis.edu, pin-yu.chen@ibm.com, yijinfeng@jd.com,}
\texttt{ecezhang@ucdavis.edu, chohsieh@ucdavis.edu}}
} 
\date{}
\usepackage{geometry}
\begin{document}

\maketitle
\begin{abstract}
We study the problem of attacking a machine learning model in the \textit{hard-label black-box} setting, where no model information is revealed except that the attacker can make queries to probe the corresponding hard-label decisions. This is a very challenging problem since the direct extension of state-of-the-art white-box attacks (e.g., C\&W or PGD) to the hard-label black-box setting will require minimizing a non-continuous step function, which is combinatorial and cannot be solved by a gradient-based optimizer. The only current approach is based on random walk on the boundary~\cite{brendel2017decision}, which requires lots of queries and lacks convergence guarantees. 
We propose a novel way to formulate the hard-label black-box attack as a real-valued optimization problem which is usually continuous and can be solved by any zeroth order optimization algorithm. For example, using the Randomized Gradient-Free method~\cite{nesterov2017random}, we are able to bound the number of iterations needed for our algorithm to achieve stationary points. We demonstrate that our proposed method outperforms the previous random walk approach to attacking convolutional neural networks on MNIST, CIFAR, and ImageNet datasets. More interestingly, we show that the proposed algorithm can also be used to attack other discrete and non-continuous machine learning models, such as Gradient Boosting Decision Trees (GBDT). 

%Almost all the previous works consider either white-box setting (model is known to the attacker) or scoring-based black-box setting (attacker can query to get model's probability output). 

\end{abstract}
\section{Introduction}
\label{sec:intro}
It has been observed recently that machine learning algorithms, especially deep neural networks, are vulnerable to adversarial examples~\cite{goodfellow2014explaining,szegedy2013intriguing,moosavi2017universal,moosavi2016deepfool,DBLP:Hongge,DBLP:journals/corr/abs-1803-01128}. For example, in image classification problems, attack algorithms~\cite{carlini2017towards,goodfellow2014explaining,chen2017zoo} can find adversarial examples for almost every image with very small human-imperceptible perturbation. 
The problem of finding an adversarial example can be posed as solving an optimization problem---within a small neighbourhood around the original example, find a point to optimize the cost function measuring the ``successfulness'' of an attack. Solving this objective function with gradient-based optimizer leads to state-of-the-art attacks \cite{carlini2017towards,goodfellow2014explaining,chen2017zoo,szegedy2013intriguing,madry2017towards}. 
%(see Section \ref{sec:related} for details). 

Most current attacks~\cite{goodfellow2014explaining,carlini2017towards,szegedy2013intriguing,chen2017ead} consider the ``white-box'' setting, where the machine learning model is fully exposed to the attacker. In this setting, the  gradient of the above-mentioned attack objective function can be computed by back-propagation, so attacks can be done very easily. 
This white-box setting is clearly unrealistic when the model parameters are unknown to an attacker. Instead, several recent works consider the ``score-based black-box'' setting, where the machine learning model is unknown to the attacker, but it is possible to make queries to obtain the corresponding probability outputs of the model~\cite{chen2017zoo,ilyas2017query}. However, in many cases real-world models will not provide probability outputs to users. Instead, only the final decision (e.g., top-1 predicted class) can be observed. It is therefore interesting to show whether machine learning model is vulnerable in this setting. 

Furthermore, existing gradient-based attacks cannot be applied to some non-continuous machine learning models which involve discrete decisions. For example, the robustness of decision-tree based models (random forest and gradient boosting decision trees (GBDT)) cannot be evaluated using gradient-based approaches, since the gradient of these functions does not exist. 

In this paper, we develop an optimization-based framework for attacking machine learning models in a more realistic and general ``hard-label black-box'' setting. We assume that the model is not revealed and the attacker can only make queries to get the corresponding {\bf hard-label decision} instead of the probability outputs (also known as soft labels). Attacking in this setting is very challenging and almost all the previous attacks fail due to the following two reasons. First, the gradient cannot be computed directly by backpropagation, and finite differences based approaches also fail because the hard-label output is insensitive to small input perturbations; second, since only hard-label decision is observed, the attack objective functions become discontinuous with discrete outputs, which is combinatorial in nature and hard to optimize (see Section \ref{sec:difficulty} for more details). 
%Currently, the only attack algorithm in this setting is based on random walk~\cite{brendel2017decision}. It requires lot of queries and is not guaranteed to converge to a meaningful solution and the convergence rate is also unknown. 

In this paper, we make hard-label black-box attacks possible and query-efficient by reformulating the attack as a novel real-valued optimization problem, which is usually continuous and much easier to solve. Although the objective function of this reformulation cannot be written in an analytical form, we show how to use model queries to evaluate its function value and apply any zeroth order optimization algorithm to solve it. Furthermore, we prove that by carefully controlling the numerical accuracy of function evaluations, a Random Gradient-Free (RGF) method can convergence to stationary points as long as the boundary is smooth. We note that this is the first attack with a guaranteed convergence rate in the hard-label black-box setting.
In the experiments, we show our algorithm can be successfully used to attack hard-label black-box CNN models on MNIST, CIFAR, and ImageNet with far less number of queries compared to the state-of-art algorithm.
%that when attacking the problem by XXX method, it is XXX times faster than decision-based attack on XXX dataset. {\color{red}(summarize the experiments)}.  

Moreover, since our algorithm does not depend on the gradient of the classifier, we can apply our approach to other non-differentiable classifiers besides neural networks. We show an interesting application in attacking Gradient Boosting Decision Tree, which cannot be attacked by all the existing gradient-based methods even in the white-box setting. Our method can successfully find adversarial examples with imperceptible perturbations for a GBDT within 30,000 queries.

\section{Background and Related work}
\label{sec:related}

We will first introduce our problem setting and give a brief literature review to hightlight the difficulty of
attacking hard-label black-box models. 

\subsection{Problem Setting}

For simplicity, we consider attacking a $K$-way multi-class classification model in this paper. 
Given the classification model $f: \R^d \rightarrow \{1, \dots, K\}$ and an original example $\bx_0$, the goal
is to generate an adversarial example $\bx$ such that 
\begin{equation}
\bx \text{ is close to } \bx_0 \ \ \ \text{ and } \ \ \ f(\bx) \neq f(\bx_0) \enskip \text{($\bx$ is misclassified by model $f$.)}
\label{eq:objective}
\end{equation}
\subsection{White-box attacks}

Most attack algorithms in the literature consider the white-box setting, where the classifier $f$ is exposed to the attacker. For neural networks, under this assumption, back-propagation can be conducted on the target model because both network structure and weights are known by the attacker.
For classification models in neural networks, it is usually assumed that $f(\bx)= \argmax_i(Z(\bx)_i)$, where $Z(\bx)\in\R^K$ is the final (logit) layer output, and $Z(\bx)_i$ is the prediction score for the $i$-th class. 
The objectives in \eqref{eq:objective} can then be naturally formulated as the following optimization problem: 
\begin{equation}
   \argmin_{\bx} \ \{ \text{Dis}(\bx, \bx_0) + c \L(Z(\bx)) \}:=h(\bx), 
   \label{eq:attack}
\end{equation}
where $\text{Dis}(\cdot, \cdot)$ is some distance measurement (e.g., $\ell_2, \ell_1$ or $\ell_\infty$ norm in Euclidean space), 
$\L(\cdot)$ is the loss function corresponding to the goal of the attack, and $c$ is a balancing parameter. 
For {\it untargeted attack}, where the goal is to make the target classifier misclassify, the loss function can be defined as
\begin{equation}
\L(Z(\bx)) = \max\{[Z(\bx)]_{y_0}-\max_{i\neq y_0}[Z(\bx)]_i,-\kappa\},
\label{eq:untargeted_loss}
\end{equation}
where $y_0$ is the original label predicted by the classifier. For {\it targeted attack}, where the goal is to turn it into a specific target class $t$, the loss function can also be defined accordingly. 

Therefore, attacking a machine learning model can be posed as solving this optimization problem~\cite{carlini2017towards,chen2017ead}, which is also known as the C\&W attack or the EAD attack depending on the choice of the distance measurement. 
To solve~\eqref{eq:attack}, one can apply any gradient-based optimization algorithm such as SGD or Adam, since the gradient of $\L(Z(\bx))$ can be computed via back-propagation. 

The ability of computing gradient also enables many different attacks in the white-box setting. For example, 
eq~\eqref{eq:attack} can also be turned into a constrained optimization problem, which can then be solved by projected gradient descent (PGD)~\cite{madry2017towards}. FGSM~\cite{goodfellow2014explaining} is the special case of one step PGD with $\ell_\infty$ norm distance. Other algorithms such as Deepfool~\cite{moosavi2016deepfool} also solve similar optimization problems to construct adversarial examples. 

\subsection{Previous work on black-box attack}

In real-world systems, usually the underlying machine learning model will not be revealed and thus
white-box attacks cannot be applied. 
This motivates the study of attacking machine learning models in the {\it black-box setting}, where attackers do not have any information about the function $f$. And the only valid operation is to make queries to the model and get the corresponding output $f(\bx)$. The first approach for black-box attack is using transfer attack~\cite{papernot2017practical}---instead of attacking the original model $f$, attackers try to construct a substitute model $\hat{f}$ to mimic $f$ and then attack $\hat{f}$ using white-box attack methods. This approach has been well studied and analyzed in~\cite{bhagoji2017exploring}. 
However, recent papers have  shown that attacking the substitute model usually leads to much larger distortion and low success rate~\cite{chen2017zoo}. Therefore, instead, \cite{chen2017zoo} considers the {\it score-based} black-box setting, where attackers can use $\bx$ to query 
the softmax layer output in addition to the final classification result. In this case, they can reconstruct the loss function \eqref{eq:untargeted_loss} and evaluate it as long as the objective function $h(\bx)$ exists for any $\bx$. Thus a zeroth order optimization approach can be directly applied to minimize $h(\bx)$. \cite{DBLP:journals/corr/abs-1805-11770} further improves the query complexity of~\cite{chen2017zoo} by introducing two novel building blocks: (i) an adaptive random gradient estimation algorithm that balances query counts and distortion, and (ii) a well-trained autoencoder that achieves attack acceleration. \cite{ilyas2017query} also solves a score-based attack problem %(assume output probability is observed) 
using an evolutionary algorithm and it shows their method could be applied to hard-label black-box setting as well. 

\subsection{Difficulty of hard-label black-box attacks}
\label{sec:difficulty}

Throughout this paper, the hard-label black-box setting refers to cases where real-world ML systems only provide limited prediction results of an input query. Specifically, only the final decision (top-1 predicted label)  instead of probability outputs is known to an attacker. 

%Unfortunately, in most realistics settings ML system will only output the final decision instead of the logit-layer scores (only $f(\bx)$ is observed instead of $Z(\bx)$). We call these XXX setting. 

\begin{figure*}[htbp]
    \centering
    \begin{tabular}{cccc}
    \subfloat[Decision boundary of $f(\bx)$\label{fig:FBN}]{\includegraphics[width=0.20\textwidth]{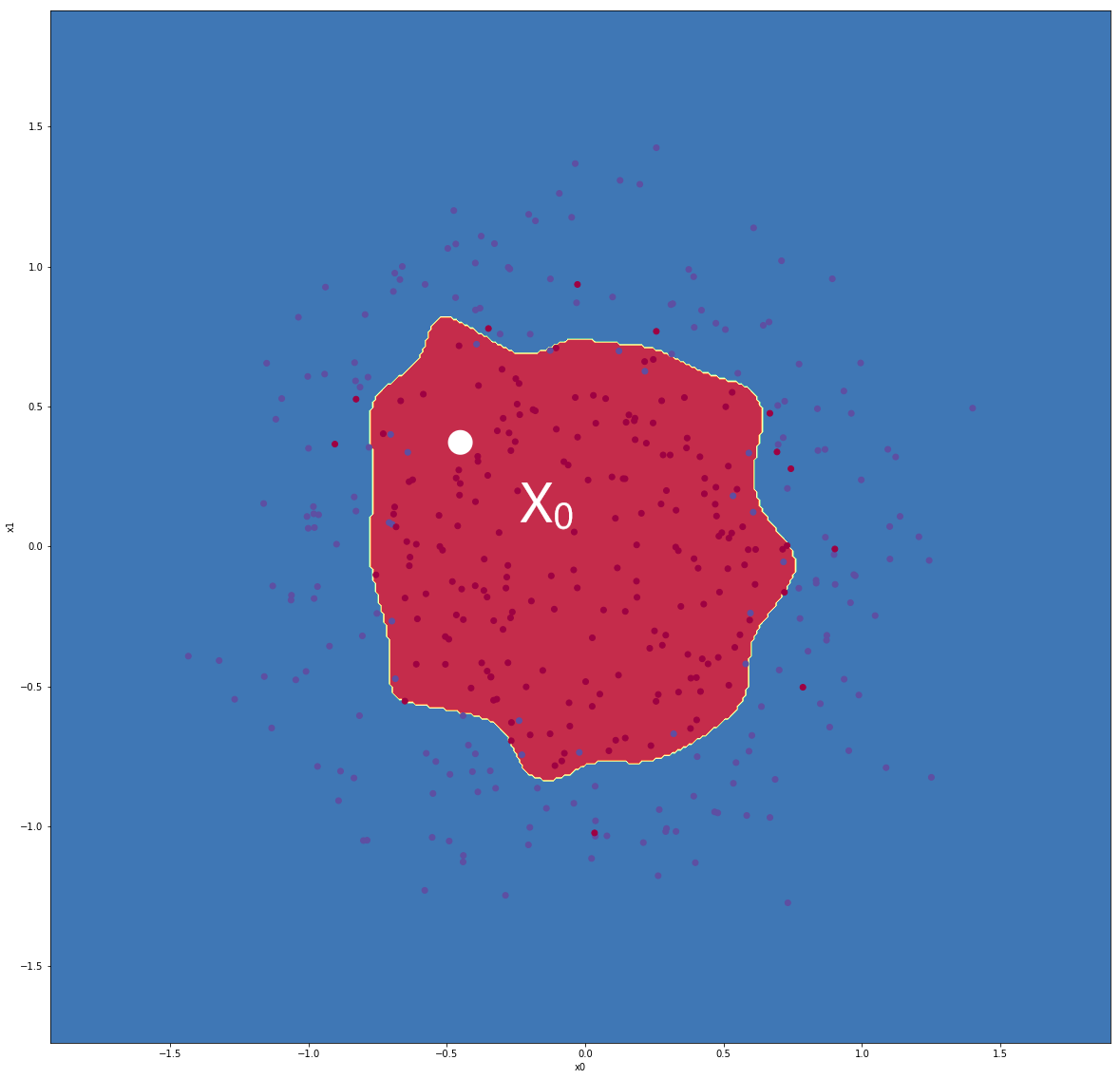}}
		&
		\subfloat[$\L(Z(\bx))$\label{fig:LZ}]{\includegraphics[width=0.22\textwidth]{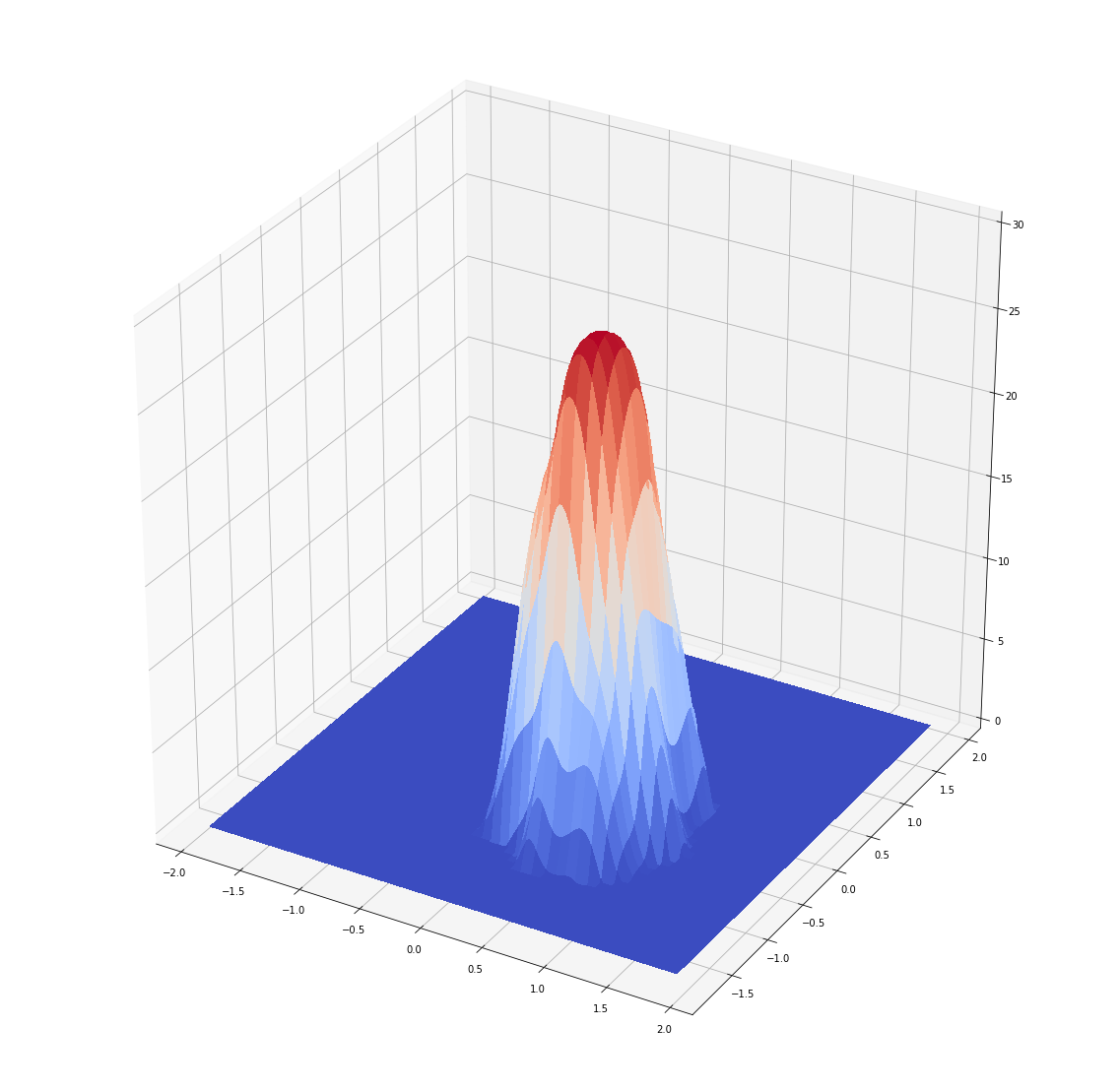}}
		&
		\subfloat[$\L(f(\bx))$\label{fig:Lf}]{\includegraphics[width=0.22\textwidth]{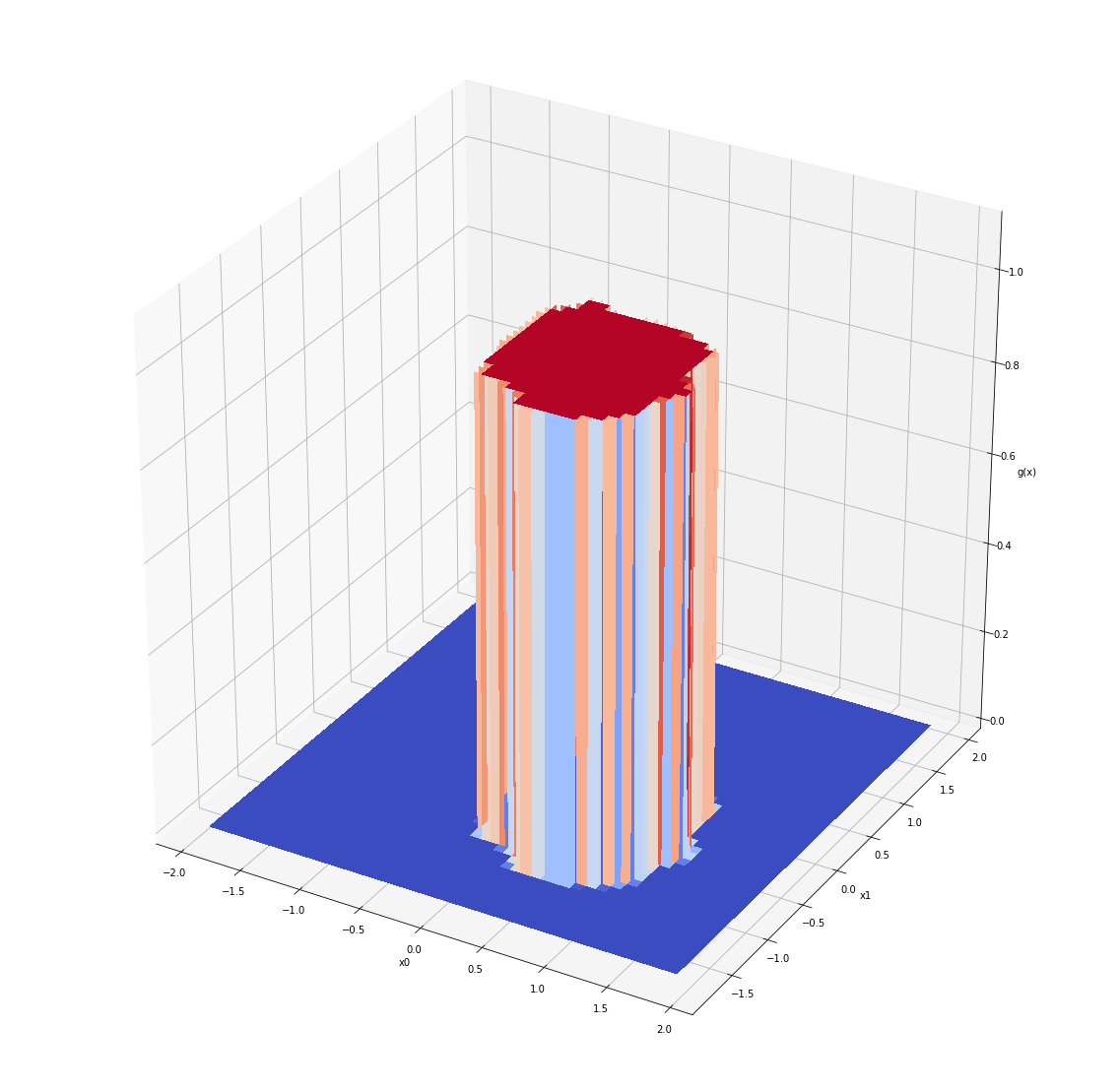}}
		&
        \subfloat[$g(\btheta)$\label{fig:GBN}]{\includegraphics[width=0.22\textwidth]{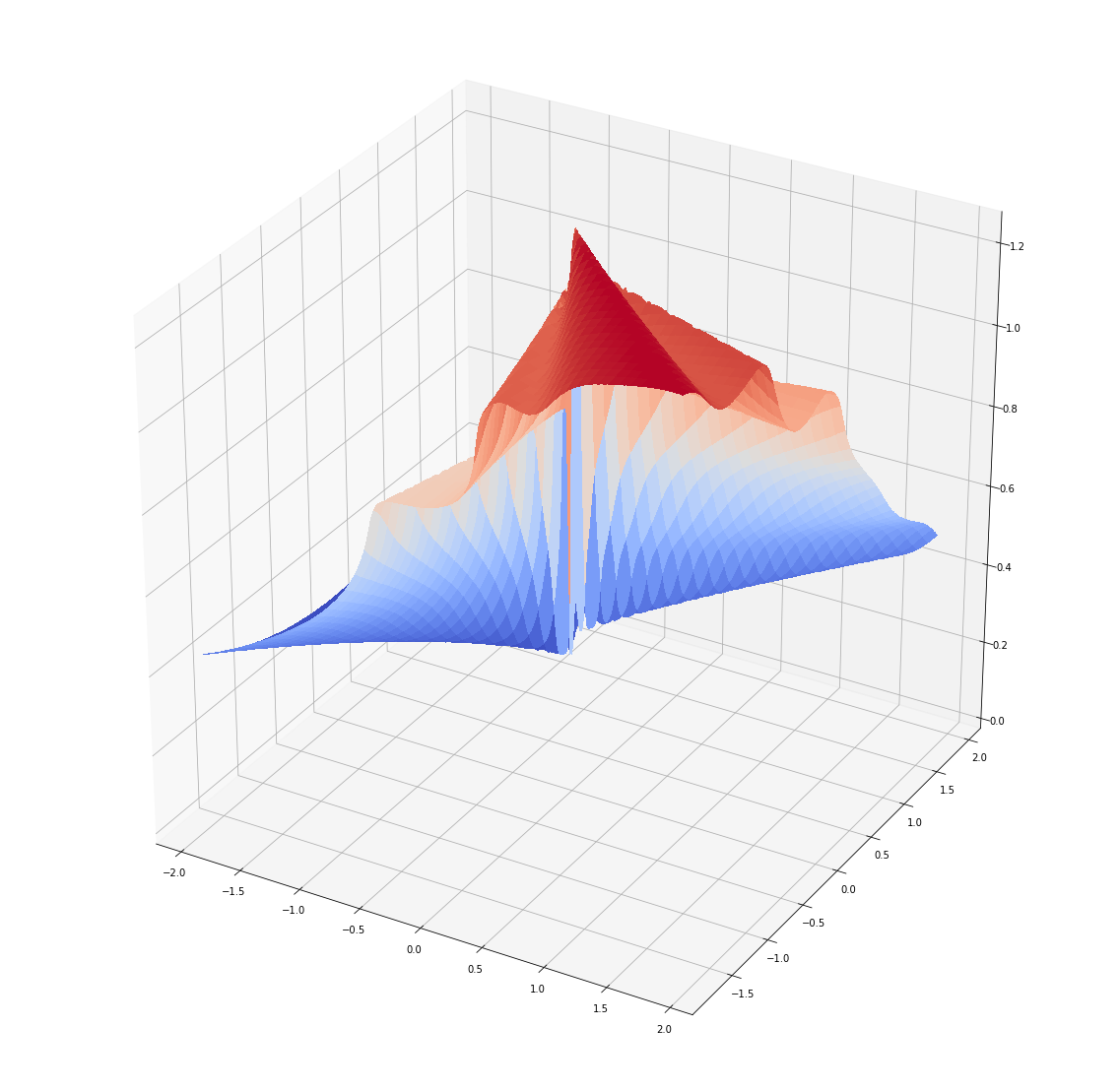}}
        \\ 
	\end{tabular}
		\caption{(a) A neural network classifier. (b) illustrates the loss function of C\&W attack, which is continuous and hence can be easily optimized. (c) is the C\&W loss function in the hard-label setting, which is discrete and discontinuous. (d) our proposed attack objective $g(\btheta)$ for this problem, which is continuous and easier to optimize. 
        %$g(\btheta)$ defines the distance from $\bx_0$ to the decision boundary along $\btheta$, 
        See detailed discussions in Section 3.}
		\label{fig:l}
\end{figure*}

Attacking in this setting is very challenging. In Figure~\ref{fig:FBN}, we show a simple 3-layer neural network's decision boundary. Note that the $\L(Z(\bx))$ term is continuous as in Figure~\ref{fig:LZ} because the logit layer output is real-valued functions. However, in the hard-label black-box setting, only $f(\cdot)$ is available instead of $Z(\cdot)$. Since $f(\cdot)$ can only be one-hot vector, if we plug-in $f$ into the loss function, $\L(f(\bx))$ (as shown in Figure \ref{fig:Lf}) will be discontinuous and with discrete outputs. 
%{\color{red} Show and plot an example here}. 

Optimizing this function will require combinatorial optimization or search algorithms, which is almost impossible to do given high dimensionality of the problem. Therefore, almost no algorithm can successfully conduct hard-label black-box attack in the literature. The only current approach~\cite{brendel2017decision} is based on random-walk on the boundary. Although this decision-based attack can find adversarial examples with comparable distortion with white-box attacks, it suffers from exponential search time, resulting in lots of queries, and lacks convergence guarantees. We show that our optimization-based algorithm can significantly reduce the number of queries compared with decision-based attack, and has guaranteed convergence in the number of iterations (queries). 
%do not have any theoretical guarantee of convergence. 

\section{Algorithms}
\label{sec:algo}

%Let $(x_0, y_0)$ denote an image $x_0$ and its true label $y_0$, and let $(x,y)$ denote the adversarial example of $x_0$ and the label $y \neq y_0$.

%The basic intuition behind the boundary attack algorithm is depicted in Figure 2: the algorithm is initialized from a point that is already adversarial and then performs a random walk along the boundary between the adversarial and the non-adversarial region such that (1) it stays in the adversarial region and (2) the distance towards the target image is reduced. In other words we perform rejection sampling with a suitable proposal distribution P to find progressively smaller adversarial perturbations according to a given adversarial criterion c(.). The basic logic of the algorithm is described in Algorithm 1, each individual building block is detailed in the next subsections

Now we will introduce a novel way to re-formulate hard-label black-box attack as another optimization problem, show how to evaluate the function value using hard-label queries, and then apply a zeroth order optimization algorithm to solve it.  

\subsection{A Boundary-based Re-formulation}

For a given example $\bx_0$, true label $y_0$ and the hard-label black-box function $f: \R^d \rightarrow \{1, \dots, K\}$, we define our objective function $g: \R^d \rightarrow \R$ depending on the type of attack:
\begin{align}
\textbf{Untargeted attack: } \ &
g(\btheta) = \texttt{argmin}_{\lambda > 0} \left( f(\bx_0 + \lambda \frac{\btheta}{\|\btheta\|}) \neq y_0 \right) \label{eq:obj_untargeted}\\
\textbf{Targeted attack (given target $t$): } \ &
g(\btheta) = \texttt{argmin}_{\lambda > 0} \left( f(\bx_0 + \lambda \frac{\btheta}{\|\btheta\|}) = t\right). \label{eq:obj_targeted}
\end{align}
%\begin{itemize}
%\item \textbf{untargeted attack:}
%	\[
%    g(\btheta) = \texttt{argmin}_{\lambda > 0} \left( f(\bx_0 + \lambda \frac{\btheta}{\|\btheta\|}) \neq y_0 \right)
%    \]
%\item \textbf{targeted attack} (given target $t$):
%	\[
%    g(\btheta) = \texttt{argmin}_{\lambda > 0} \left( f(\bx_0 + \lambda \frac{\btheta}{\|\btheta\|}) = t\right)
%    \]
%\end{itemize}

In this formulation, $\btheta$ represents the search direction 
%pointing from $\bx_0$ to the decision boundary, 
and $g(\btheta)$ is the distance  from $\bx_0$ to the nearest adversarial example along the direction $\btheta$. The difference between \eqref{eq:obj_untargeted} and \eqref{eq:obj_targeted} corresponds to the different definitions of ``successfulness'' in untargeted and targeted attack, where the former one aims to turn the prediction into any incorrect label and the later one aims to turn the prediction into the target label. 
For untargeted attack, $g(\btheta)$ also corresponds to the distance to  the decision boundary along the direction $\btheta$. 
In image problems the input domain of $f$ is bounded, so we will add corresponding upper/lower bounds in the definition of \eqref{eq:obj_untargeted} and \eqref{eq:obj_targeted}. 
\\
\begin{wrapfigure}{r}{0.3\textwidth} %this figure will be at the right
    \centering
    \includegraphics[width=0.3\textwidth]{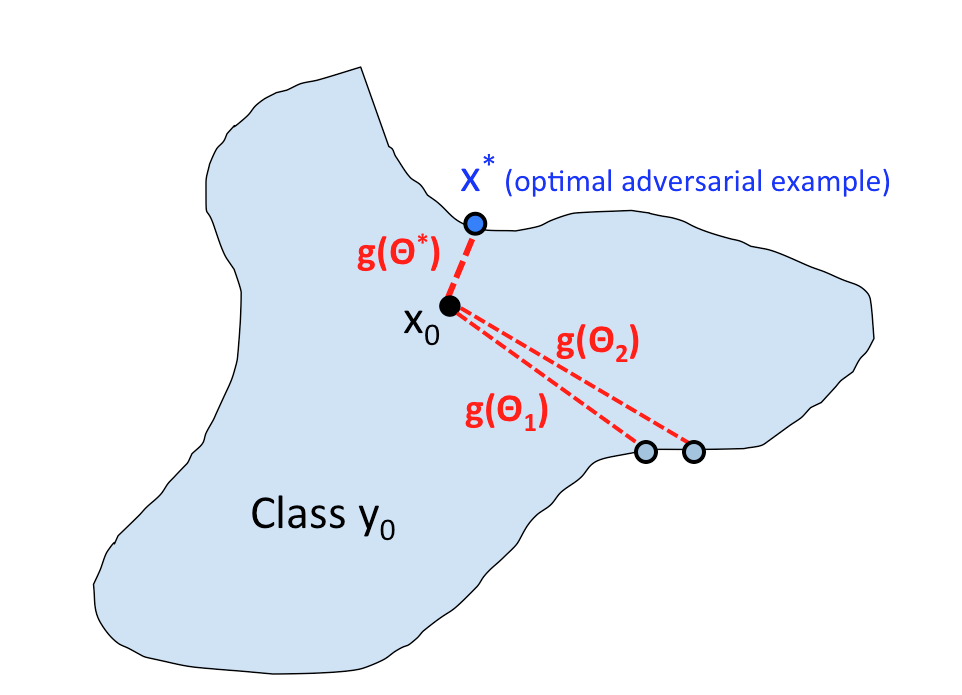}
    \caption{Illustration}
    \label{fig:illustration}
\end{wrapfigure}
Instead of searching for an adversarial example, we search the direction $\btheta$ to minimize the distortion $g(\btheta)$, which leads to the following optimization problem:
%The problem of finding an adversarial example can be formulated as an optimization problem,
\begin{equation}
\label{equ:obj}
\min\limits_{\btheta} \  g(\btheta).
\end{equation}
Finally, the adversarial example can be found by $\bx^* = \bx_0 + g(\btheta^*)\frac{\btheta^*}{\|\btheta^*\|}$, where $\btheta^*$ is the optimal solution of (\ref{equ:obj}). 

Note that unlike the C\&W or PGD objective functions, which are discontinuous step functions in the hard-label setting (see Section~\ref{sec:related}), $g(\btheta)$ maps input direction to real-valued output (distance to decision boundary), which is usually continuous---a small change of $\btheta$ usually leads to a small change of $g(\btheta)$, as can be seen from Figure~\ref{fig:illustration}. 

\begin{figure*}[tbp]
    \centering
    \begin{tabular}{cccc}
    \subfloat[Decision boundary of continuous function\label{fig:FB1}]{\includegraphics[width=0.20\textwidth]{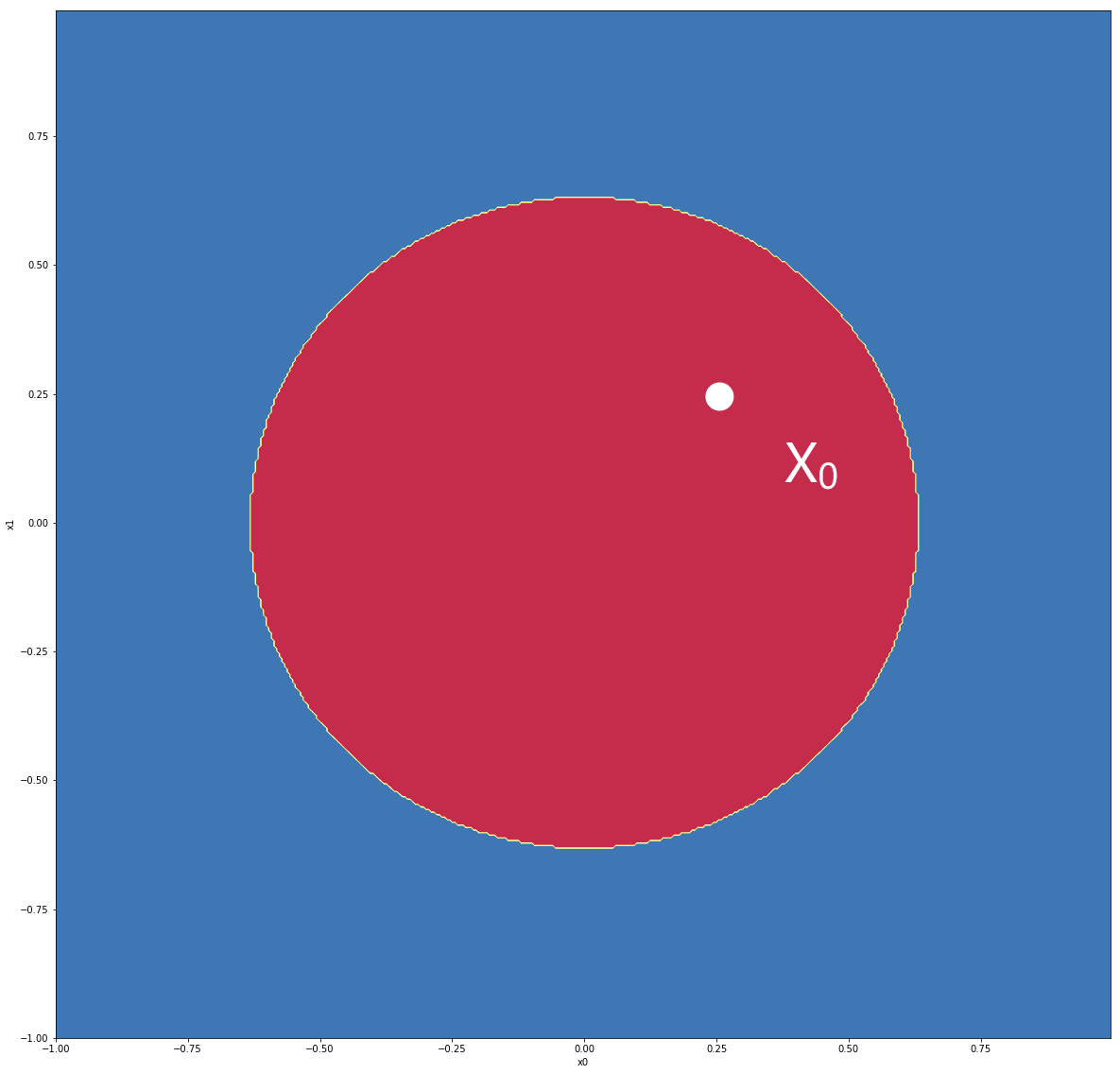}}
		&
		\subfloat[Decision boundary of GBDT\label{fig:FB2}]{\includegraphics[width=0.20\textwidth]{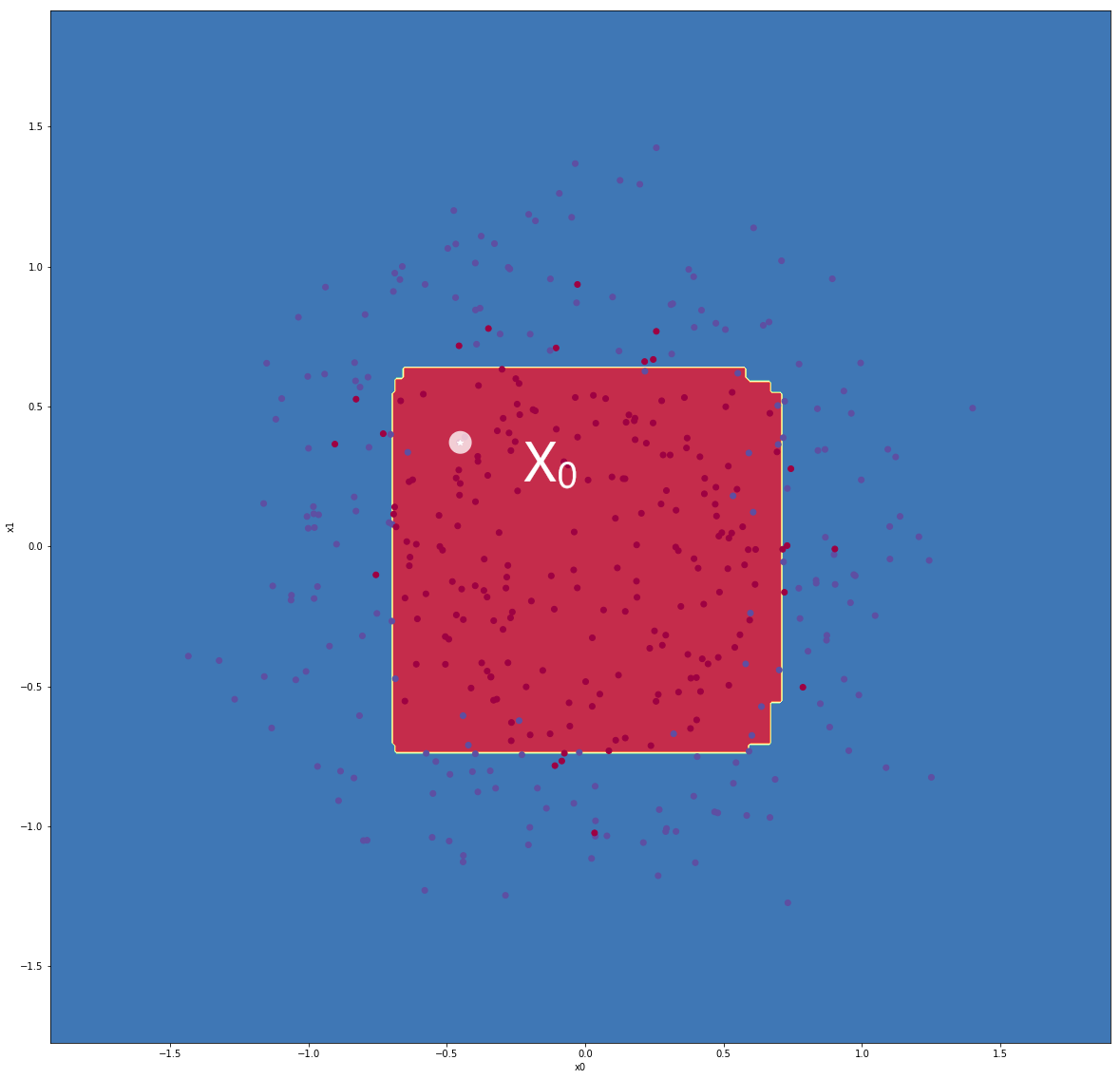}}
		&
         \subfloat[$g(\btheta)$ of (a)\label{fig:GB1}]{\includegraphics[width=0.22\textwidth]{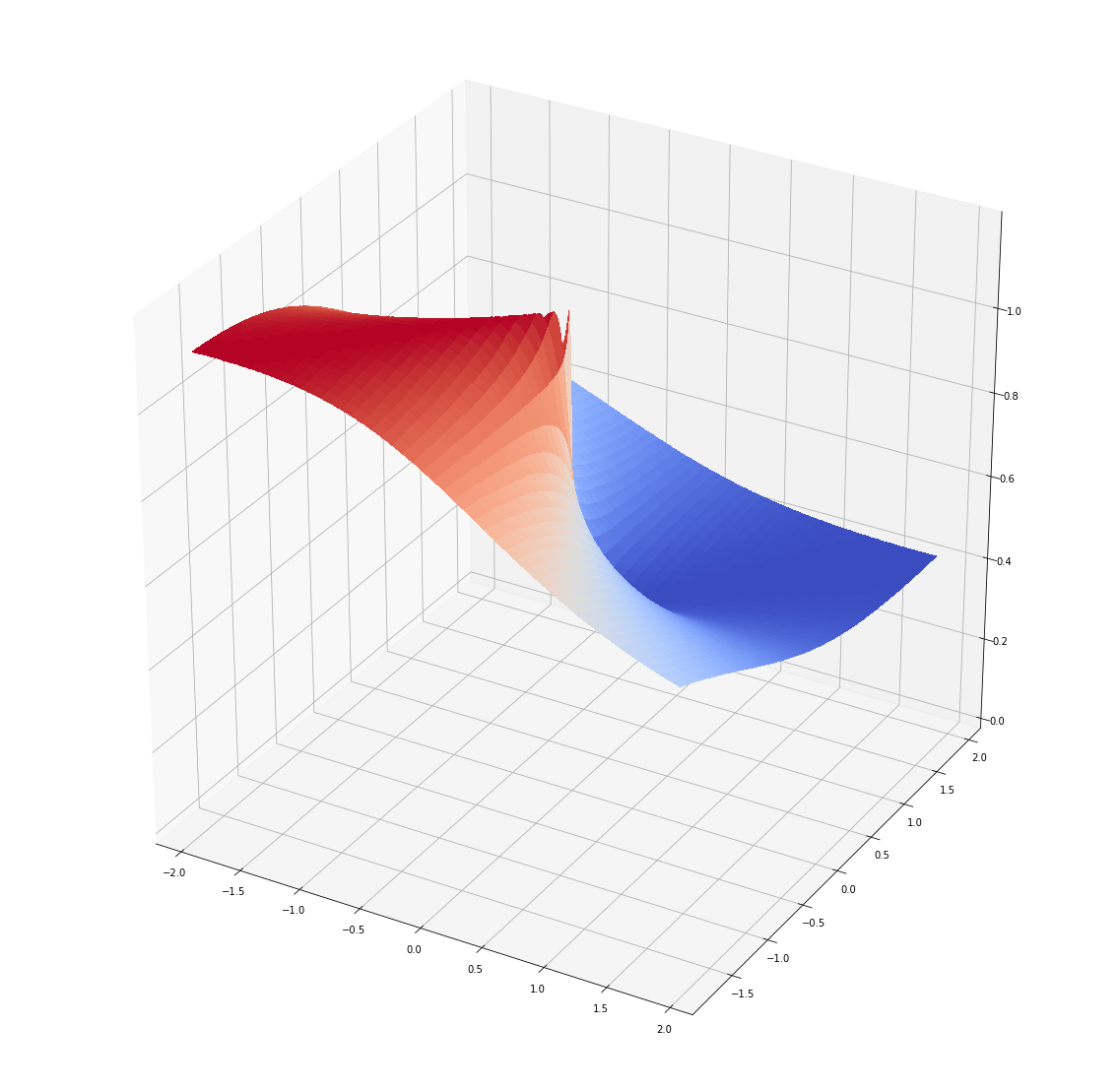}}
		&
		\subfloat[$g(\btheta)$ of (b)\label{fig:GB2}]{\includegraphics[width=0.22\textwidth]{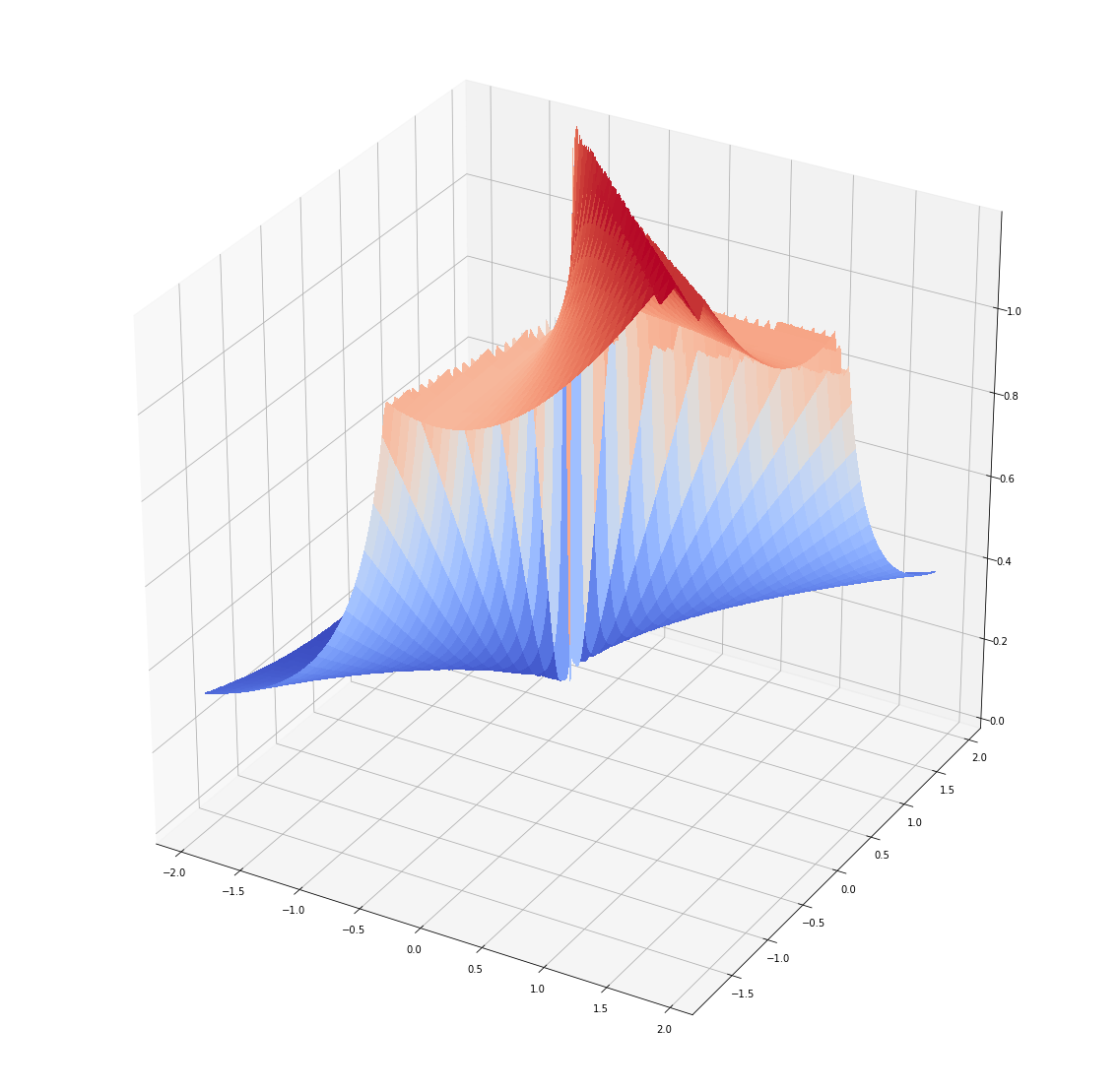}}\\
	\end{tabular}
		\caption{Examples of decision boundary of classification function $f(\bx)$ and corresponding $g(\btheta)$ }
		\label{fig:g}
\end{figure*}

Moreover, we give three examples of $f(\bx)$ defined in two dimension input space and their corresponding $g(\btheta)$. In Figure ~\ref{fig:FB1}, we have a continuous classification function defined as follows \[
    f(\bx)= 
\begin{cases}
    1,& \text{if } \|\bx\|^2\geq 0.4\\
    0,              & \text{otherwise.}
\end{cases}
\] In this case, as shown in Figure ~\ref{fig:GB1}, $g(\btheta)$ is continuous. Moreover, in Figure ~\ref{fig:FB2} and Figure ~\ref{fig:FBN}, we show decision boundaries generated by GBDT and neural network classifier, which are not continuous. However, as showed in Figure ~\ref{fig:GB2} and Figure ~\ref{fig:GBN}, even if the classifier function is not continuous, $g(\btheta)$ is still continuous. This makes it easy to apply zeroth order method to solve~\eqref{equ:obj}.
%Therefore, we are able to apply zeroth order optimization techniques to solve (\ref{equ:obj}).  

{\bf Compute $g(\btheta)$ up to certain accuracy. }
We are not able to evaluate the gradient of $g$, but we can evaluate the function value of $g$ using the hard-label queries to the original function $f$. For simplicity, we focus on untargeted attack here, but the same procedure can be applied to targeted attack as well. 

First, we discuss how to compute $g(\btheta)$ directly without additional information. This is used in the initialization step of our algorithm. For a given normalized $\btheta$, we do a fine-grained search and then a binary search. In fine-grained search, we query the points $\{\bx_0+\alpha\btheta, \bx_0+2\alpha\btheta, \dots\}$ one by one until we find $f(\bx+ i\alpha\btheta)\neq y_0$. This means the boundary goes between $[\bx_0+(i-1)\alpha\btheta, \bx_0 + i\alpha \btheta]$. We then enter the second phase and conduct a binary search to find the solution within this region (same with line 11--17 in Algorithm~\ref{alg:compute_g}). Note that there is an upper bound of the first stage if we choose $\btheta$ by the direction of $\bx-\bx_0$ with some $\bx$ from another class. 
%Even for a random direction, in image applications the data is naturally bounded, so an upper bound still exists. 
This procedure is used to find the initial $\btheta_0$ and corresponding $g(\btheta_0)$ in our optimization algorithm. We omit the detailed algorithm for this part since it is similar to Algorithm~\ref{alg:compute_g}. 
%In practice we never reach the upper bound in all our experiments.

Next, we discuss how to compute $g(\btheta)$ when we know the solution is very close to a value $v$. This is used in all the function evaluations in our optimization algorithm, since the current solution is usually close to the previous solution, and when we estimate the gradient using~\eqref{eq:g_estimate}, the queried direction will only be a small perturbation of the previous one. In this case, we first increase or decrease $v$ in local region to find the interval that contains boundary (e.g, $f(v)=y_0$ and $f(v')\neq y_0$), then conduct a binary search to find the final value of $g$. Our procedure for computing $g$ value is presented in Algorithm~\ref{alg:compute_g}. 

\begin{algorithm}
\caption{Compute $g(\btheta)$ locally}\label{alg:compute_g}
\begin{algorithmic}[1]
%\Procedure{Euclid}{$a,b$}\Comment{The g.c.d. of a and b}
\State \textbf{Input: } Hard-label model $f$, original image $x_0$, query direction $\btheta$, previous value $v$, increase/decrease ratio $\alpha=0.01$, stopping tolerance $\epsilon$ (maximum tolerance of computed error)
\State $\btheta \leftarrow \btheta/\|\btheta\|$
\If{$f(\bx_0+v\btheta)= y_0$}
 \State $v_{left}\leftarrow v, v_{right}\leftarrow (1+\alpha) v$
 \While{$f(\bx_0+v_{right}\btheta)=y_0$}
     \State $v_{right} \leftarrow (1+\alpha) v_{right}$
 \EndWhile
\Else
\State $v_{right}\leftarrow v, v_{left} \leftarrow (1-\alpha)v$
\While{$f(\bx_0+v_{left}\btheta)\neq y_0$}
\State $v_{left}\leftarrow (1-\alpha) v_{left}$
\EndWhile
\EndIf
\State \#\# Binary Search within $[v_{left}, v_{right}]$
\While{$v_{right}-v_{left} > \epsilon$}
    \State $v_{mid} \leftarrow (v_{right}+v_{left})/2$
    \If{$f(\bx_0+v_{mid}\btheta) = y_0$}
        \State $v_{left} \leftarrow v_{mid} $
        \Else
        \State $v_{right} \leftarrow v_{mid}$
    \EndIf
\EndWhile
%\While{ query coun $\le Q$} 
%\State Set $\btheta = \bx_i - \bx_0$ where $\bx_i$ is the closet point to $\bx_0$
\State \textbf{return} $v_{right}$ %\Comment{The gcd is b}
%\EndProcedure
\end{algorithmic}
\end{algorithm}

\subsection{Zeroth Order Optimization}

To solve the optimization problem~\eqref{eq:objective} for which we can only evaluate function value instead of gradient, zeroth order optimization algorithms can be naturally applied. In fact, after the reformulation, the problem can be potentially solved by any zeroth order optimization algorithm, like zeroth order gradient descent or coordinate descent (see~\cite{conn2009introduction} for a comprehensive survey). 

Here we propose to solve~\eqref{eq:objective} using Randomized Gradient-Free (RGF) method proposed in~\cite{nesterov2017random,ghadimi2013stochastic}. In practice we found it outperforms zeroth-order coordinate descent. In each iteration, the gradient is estimated by 
\begin{equation}
\hat{\bg} =  \dfrac{g(\btheta + \beta \bu) - g(\btheta)}{\beta} \cdot \bu 
\label{eq:g_estimate}
\end{equation}
where $\bu$ is a random Gaussian vector, and $\beta > 0$ is a smoothing parameter (we set $\beta = 0.005$ in all our experiments).
The solution is then updated by $\btheta\leftarrow \btheta - \eta \hat{\bg}$ with a step size $\eta$. The procedure is summarized in Algorithm~\ref{alg:algo}.
%The hyperparameter $b$ is a tunable parameter that balances the bias and variance trade-off of the gradient estimation error.

\begin{algorithm}
\caption{RGF for hard-label black-box attack}\label{alg:algo}
\begin{algorithmic}[1]
%\Procedure{Euclid}{$a,b$}\Comment{The g.c.d. of a and b}
\State \textbf{Input: } Hard-label model $f$, original image $x_0$, initial $\btheta_0$.
%\State Set $\btheta = \bx_i - \bx_0$ where $\bx_i$ is the closet point to $\bx_0$
\For{$t=0, 1, 2, \ldots, T$}
%\While{ query count $\le Q$} %\Comment{We have the answer if r is 0}
\State Randomly choose $\bu_t$ from a zero-mean Gaussian distribution
\State Evaluate $g(\btheta_t)$ and $g(\btheta_t+\beta \bu)$ using Algorithm~\ref{alg:compute_g}
\State Compute \quad $\hat{\bg} = \dfrac{g(\btheta_t + \beta \bu) - g(\btheta_t)}{\beta} \cdot \bu$
\State Update \quad $\btheta_{t+1} = \btheta_t - \eta_t \hat{\bg}$
\EndFor\label{euclideanwhile}
%\EndWhile\label{euclidendwhile}
\State \textbf{return} $\bx_0 +  g(\btheta_T)\btheta_T$ %\Comment{The gcd is b}
%\EndProcedure
\end{algorithmic}
\end{algorithm}
There are several implementation details when we apply this algorithm. First, for high-dimensional problems, we found the estimation in~\eqref{eq:g_estimate} is very noisy. Therefore, instead of using one vector, we sample $q$ vectors from Gaussian distribution and average their estimators to get $\hat{\bg}$. We set $q=20$ in all the experiments. The convergence proofs can be naturally extended to this case. 
Second, instead of using a fixed step size (suggested in theory), we use a backtracking line-search approach to find step size at each step. This leads to additional query counts, but makes the algorithm more stable and eliminates the need to hand-tuning the step size. 

\subsection{Theoretical Analysis}

\label{sec:convergence}
If $g(\btheta)$ can be computed exactly, it has been proved in~\cite{nesterov2017random} that RGF in Algorithm~\ref{alg:algo} requires at most $O(\frac{d}{\delta^2})$ iterations to converge to a point with $\|\nabla g(\btheta)\|^2 \leq \delta^2$. 
However, in our algorithm the function value $g(\btheta)$ cannot be computed exactly; instead, we compute it up to $\epsilon$-precision, and this precision can be controlled by binary threshold in Algorithm~\ref{alg:compute_g}. 
We thus extend the proof in~\cite{nesterov2017random} to include the 
case of approximate function value evaluation, as described in the following theorem. 
%Now we analyze the convergence rate of the proposed algorithm. Inspired by~\cite{nesterov2011random}, we show that even if $g$ is non-convex function, with stopping criterion $\epsilon\sim O(\beta\delta^2)$, we could get $O(\frac{d}{\delta^2})$ in smooth case 
\begin{theorem}
In Algorithm~\ref{alg:algo}, 
suppose g has Lipschitz-continuous gradient with constant $L_1(g)$. If the error of function value evaluation is controlled by $\epsilon\sim O(\beta\delta^2)$ and $\beta\leq O(\frac{\delta}{dL_1(g)})$, then in order to obtain $\frac{1}{N+1}\sum\limits_{k=0}^N E_{\mathcal{U}_k}(\|\nabla g(\btheta_k)\|^2) \leq \delta^2$, the total number of iterations is at most $O(\frac{d}{\delta^2})$.
\end{theorem}
Detailed proofs can be found in the appendix. Note that the binary search procedure could obtain the desired function value precision in $O(\log\delta)$ steps. By using the same idea with Theorem 1 and following the proof in~\cite{nesterov2017random}, we could also achieve $O(\frac{d^2}{\delta^3})$ complexity when $g(\btheta)$ is non-smooth but Lipschitz continuous. 
\section{Experimental results}
\label{sec:exp}
%\subsection{Comparison of different optimization algorithms? (coordinate descent vs random direction)}

We test the performance of our hard-label black-box attack algorithm on convolutional neural network (CNN) models and compare with decision-based attack~\cite{brendel2017decision}. Furthermore, we show our method can be applied to attack Gradient Boosting Decision Tree (GBDT) and present some interesting findings. 

\subsection{Attack CNN image classification models}
We use three standard datasets: MNIST~\cite{lecun1998gradient}, CIFAR-10~\cite{krizhevsky2009learning} and ImageNet-1000~\cite{deng2009imagenet}. To have a fair comparison with previous work, we adopt the same networks used in both ~\cite{carlini2017towards} and ~\cite{brendel2017decision}. In detail, both MNIST and CIFAR use the same network structure with four convolution layers, two max-pooling layers and two fully-connected layers. Using the parameters provided by ~\cite{carlini2017towards}, we could achieve 99.5\% accuracy on MNIST and 82.5\% accuracy on CIFAR-10, which is similar to what was reported in ~\cite{carlini2017towards}. For Imagenet-1000, we use the pretrained network Resnet-50 ~\cite{he2016deep} provided by torchvision\footnote{https://github.com/pytorch/vision/tree/master/torchvision}, which could achieve 76.15\% top-1 accuracy. All models are trained using Pytorch and our source code is publicly available\footnote{https://github.com/LeMinhThong/blackbox-attack}.

We include the following algorithms into comparison: 
\begin{compactitem}
\item Opt-based black-box attack (Opt-attack): our proposed algorithm. 
%We solve the proposed optimization problem~\eqref{eq:objective} using Random Stochastic Gradient (RSG) method. 
%We choose 1000 randomly draw images from training datasets as initialization.
\item Decision-based black-box attack~\cite{brendel2017decision} (Decision-attack): the only previous work on attacking hard-label black box model. We use the authors' implementation and use default parameters provided in Foolbox\footnote{https://github.com/bethgelab/foolbox}.
\item C\&W white-box attack~\cite{carlini2017towards}: one of the current state-of-the-art attacking algorithm in the white-box setting. We do binary search on parameter $c$ per image to achieve the best performance. Attacking in the white-box setting is a much easier problem, so we include C\&W attack just for reference and indicate the best performance we can possibly achieve. 
\end{compactitem}
%The only previous work for attacking hard-label black box model is the decision-based attack proposed in~\cite{brendel2017decision}. 

For all the cases, we conduct adversarial attacks for randomly sampled $N=100$ images from validation sets. Note that all three attacks have 100\% successful rate, and we report the average $L_2$ distortion, defined by
%\begin{equation*}
   $ \frac{1}{N}\sum_{i=1}^N \|\bx^{(i)} - \bx_0^{(i)}\|_2$, 
where $\bx^{(i)}$ is the adversarial example constructed by an attack algorithm and $\bx_0^{(i)}$ is the original $i$-th example. For black-box attack algorithms, we also report average number of queries for comparison.
%for all the $N$ images. 
\subsubsection{Untargeted Attack}

%\begin{table}
%  \caption{Results of untargeted and targeted attack}
%  \label{tab:results}
%  \centering
%  \begin{tabular}{cccccc}
%    \toprule
%  %  \cmidrule(r){1-2}
%    &  \multicolumn{2}{c}{Opt-based Attack (black-box)} &\multicolumn{2}{c}{Decision-based Attack (black-box)} & C\&W (white-box)\\
%    &Avg $L_2$& \# queries& Avg $L_2$ & \# queries& Avg $L_2$ \\
%    \midrule 
%    \multirow{3}{*}{MNIST (untargeted)}  & & & & & \\
%       & & & & & \\
%          & & & & & \\
%          \hline
%   \multirow{3}{*}{MNIST (targeted)}   & & & & & \\
%       & & & & & \\
%          & & & & & \\
%          \hline
%            \multirow{3}{*}{CIFAR (untargeted)}   & & & & & \\
%       & & & & & \\
%          & & & & & \\
%          \hline
%            \multirow{3}{*}{CIFAR (targeted)}   & & & & & \\
%       & & & & & \\
%          & & & & & \\
%          \hline
%            \multirow{3}{*}{ImageNet (untargeted)}   & & & & & \\
%       & & & & & \\
%          & & & & & \\
%          \hline
%    \bottomrule
%  \end{tabular}
%\end{table}

%We have the following findings: 
%\begin{compactitem}
%\item Our algorithm consistently achieves 
%\end{compactitem}
\begin{table}
  \caption{Results of untargeted attack.}
  \label{un-table}
  \centering
  \begin{tabular}{lcccccc}
    \toprule
    \cmidrule(r){1-2}
    &  \multicolumn{2}{c}{MNIST} &\multicolumn{2}{c}{CIFAR10} & \multicolumn{2}{c}{Imagenet (ResNet-50)}\\
    &Avg $L_2$& \# queries& Avg $L_2$ & \# queries& Avg $L_2$ & \# queries\\
    \midrule 
    \multirow{2}{*}{Decision-attack (black-box)}   &1.1222 &60,293 & 0.1575 & 123,879 & 5.9791 & 123,407 \\
    &1.1087 &143,357 & 0.1501 & 220,144 & 3.7725 &  260,797\\
    \hline
    \multirow{2}{*}{Opt-attack (black-box)} &  1.188 &  22,940 &0.2050 & 40,941 & 6.9796  & 71,100 \\
   & 1.049 & 51,683 & 0.1625 & 77,327& 4.7100 & 127,086\\
   & 1.011  & 126,486 & 0.1451 & 133,662 & 3.1120 &  237,342\\
   \hline
      C\&W (white-box)      & 0.9921  & - & 0.1012 & - & 1.9365 & -   \\
        \bottomrule
  \end{tabular}
\end{table}

\begin{wrapfigure}[15]{r}{0.45\textwidth} %this figure will be at the right
    \centering
    \includegraphics[width=0.4\textwidth]{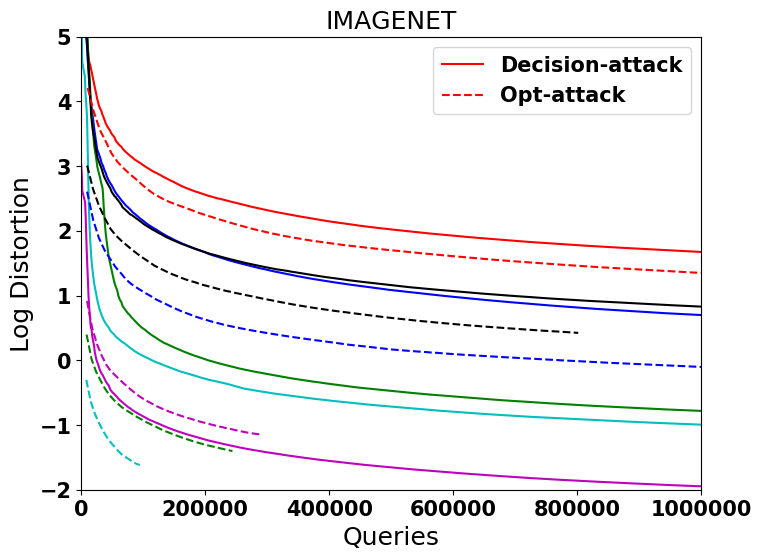}
    \caption{\small Log distortion comparison of Decision-attack (solid curves) vs Opt-attack (dotted curves) over number of queries for 6 different images. }
    \label{fig:imgnet}
\end{wrapfigure}
%The results in Table~\ref{un-table} show that 
For untargeted attack, the goal is to turn a correctly classified image into any other label. The results are presented in Table~\ref{un-table}. Note that for both Opt-attack and Decision-attack, by changing stopping conditions we can get the performance with different number of queries. 

First, we compare two black-box attack methods in Table~\ref{un-table}. Our algorithm consistently achieves smaller distortion with less number of queries than Decision-attack. For example, on MNIST data, we are able to reduce the number of queries by 3-4 folds, and Decision-attack converges to worse solutions in all the 3 datasets. Compared with C\&W attack, we found black-box attacks attain slightly worse distortion on MNIST and CIFAR. 
\begin{figure}
\centering
\begin{tabular}{c}
\subfloat[Examples of targeted Opt-attack]{\includegraphics[width=0.9\textwidth]{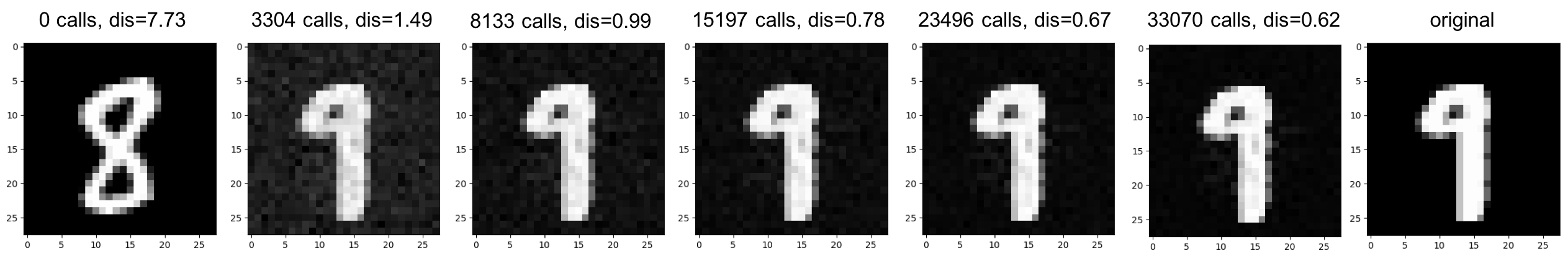}}\\
\subfloat[Examples of targeted Decision-attack]{\includegraphics[width=0.9\textwidth]{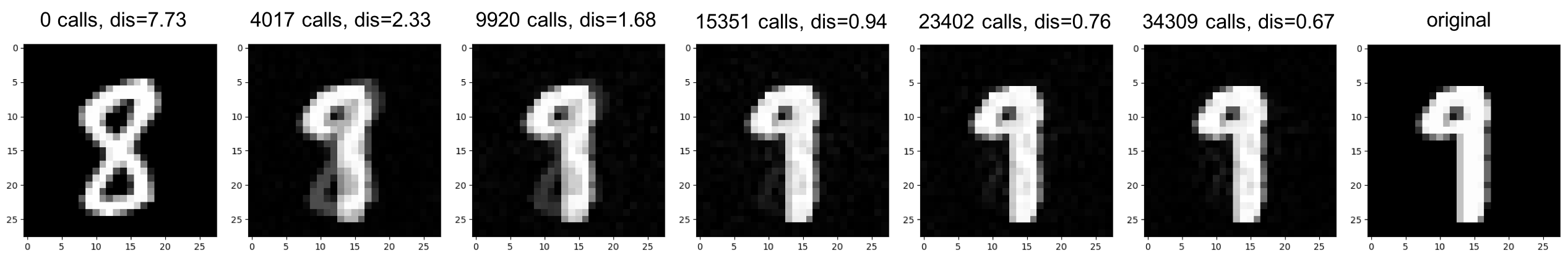}}\\ 
\end{tabular}
\caption{Example quality comparison between targeted Opt-attack  and Decision-attack. Opt-attack can achieve a better result with less queries.  }
\label{fig:examples}
\end{figure}
This is reasonable because white-box attack has much more information than black-box attack and is strictly easier. 
We note that the experiments in~\cite{brendel2017decision} conclude that C\&W and Decision-attack have similar performance because they only run C\&W with a single regularization parameter $c$ without doing binary search to obtain the optimal parameter. For ImageNet, since we constraint the number of queries, the distortion of black-box attacks is much worse than C\&W attack. The gap can be reduced by increasing the number of queries as showed in Figure ~\ref{fig:imgnet}. 

\subsubsection{Targeted attack}
The results for targeted attack is presented in Table~\ref{target-table}. Following the experiments in~\cite{brendel2017decision}, for each randomly sampled image with label $i$ we set target label $t=(i+1)\ \text{module}\ 10$. On MNIST data, we found our algorithm is more than 4 times faster (in terms of number of queries) than Decision-attack and converge to a better solution. On CIFAR data, our algorithm has similar efficiency with Decision-attack at the first 60,000 queries, but converges to a slightly worse solution. Also, we show a example quality comparison from the same starting point to the original sample in Figure~\ref{fig:examples}.

\begin{table}
  \caption{Results of targeted attack.}
  \label{target-table}
  \centering
  \begin{tabular}{lcccc}
    \toprule
 %   \cmidrule(r){1-2}
    &   \multicolumn{2}{c}{MNIST} &\multicolumn{2}{c}{CIFAR10}\\
    &Avg $L_2$& \# queries& Avg $L_2$& \# queries\\
    \midrule 
        \multirow{3}{*}{Decision-attack (black-box)}   &2.3158&30,103 & 0.2850&55,552      \\
     & 2.0052 & 58,508  &0.2213 & 140,572  \\
 & 1.8668 & 192,018   & 0.2122 & 316,791   \\
    \hline
    \multirow{3}{*}{Opt-attack (black-box)} & 1.8522 & 46,248 &0.2758 & 61,869 \\
    & 1.7744& 57,741 & 0.2369 & 141,437 \\
    & 1.7114 & 73,293 & 0.2300 & 186,753\\
    \hline
    C\&W (white-box)      &  1.4178 & - & 0.1901 & -   \\
    \bottomrule
  \end{tabular}
\end{table}
\begin{table}[t]
  \caption{Results of untargeted attack on gradient boosting decision tree.}
  \label{gbdt-table}
  \centering
  \begin{tabular}{ccccc}
    \toprule
  %  \cmidrule(r){1-2}
    &   \multicolumn{2}{c}{HIGGS} &\multicolumn{2}{c}{MNIST}\\
    &Avg $L_2$& \# queries& Avg $L_2$& \# queries\\
    \midrule
    \multirow{3}{*}{Ours} & 0.3458 & 4,229 & 0.6113 & 5,125 \\
    &0.2179 & 11,139 & 0.5576 & 11,858 \\
    & 0.1704 &29,598& 0.5505 & 32,230\\
    \bottomrule
  \end{tabular}
\end{table}

\subsubsection{Attack Gradient Boosting Decision Tree (GBDT)}
To evaluate our method's ability to attack models with discrete decision functions, we conduct our untargeted attack on gradient booting decision tree (GBDT).   
In this experiment, we use two standard datasets: HIGGS~\cite{baldi2014searching} for binary classification and MNIST~\cite{lecun1998gradient} for multi-class classification. 
We use popular LightGBM\footnote{https://github.com/Microsoft/LightGBM} framework to train the GBDT models. Using suggested parameters\footnote{https://github.com/Koziev/MNIST\_Boosting}, we could achieve 0.8457 AUC for HIGGS and 98.09\% accuracy for MNIST. The results of untargeted attack on GBDT are in Table~\ref{gbdt-table}.

As shown in Table ~\ref{gbdt-table}, by using around 30K queries, we could get a small distortion on both datasets, which firstly uncovers the vulnerability of GBDT models. Tree-based methods are well-known for its good interpretability. And because of that, they are widely used in the industry. However, we show that even with good interpretability and a similar prediction accuracy with convolution neural network, the GBDT models are vulnerable under our Opt-attack. This result raises a question about tree-based models' robustness, which will be an interesting direction in the future.

%\subsubsection{Show some pictures in MNIST}

\section{Conclusion}

In this paper, we propose a generic and optimization-based hard-label black-box attack algorithm, which can be applied to discrete and non-continuous models other than neural networks, such as the gradient boosting decision tree. Our method enjoys query-efficiency and has a theoretical convergence guarantee on the attack performance. Moreover, our attack achieves smaller or similar distortion using 3-4 times less queries compared with the state-of-the-art algorithm.

\bibliographystyle{unsrt}
\bibliography{nips_2018}
\newpage
\section{Appendix}

Because there is a stopping criterion in Algorithm ~\ref{alg:compute_g}, we couldn't achieve the exact $g(\btheta)$. Instead, we could get $\tilde{g}$ with $\epsilon$ error, i.e., $g(\btheta) - \epsilon \leq \tilde{g}(\btheta) \leq g(\btheta) + \epsilon$. Also, we define $\hat{\bg}(\btheta) = \frac{\tilde{g}(\btheta+\beta \bu)-\tilde{g}(\btheta)}{\beta} \cdot \bu$ to be the noisy gradient estimator. 

Following \cite{nesterov2011random}, we define the Guassian smoothing approximation over $g(\theta)$, i.e,
\begin{equation}
g_\beta(\theta) = \frac{1}{\kappa}\int_E g(\theta + \beta u)e^{-\frac{1}{2}||u||^2} du.
\end{equation}
Also, we have the upper bounds for the moments $M_p=\frac{1}{\kappa}\int_E ||u||^pe^{-\frac{1}{2}||u||^2} du$ from ~\cite{nesterov2011random} Lemma 1. 

For $p \in [0,2]$, we have
\begin{equation}
M_p \leq d^{p/2}. 
\end{equation}
If $p \geq 2$, we have two-sided bounds
\begin{equation}
	n^{p/2} \leq M_p \leq (p+n)^{p/2}. 
\end{equation}
\subsection{Proof of Theorem 1}
Suppose $g$ has a lipschitz-continuous gradient with constant $L_1(g)$, then
\begin{equation}
|g(y)-g(x)-\langle\nabla g(x),y-x\rangle| \leq \frac{1}{2}L_1(g)||x-y||^2.
\end{equation}
We could bound $E_u(||\hat{g}(\btheta)||^2)$ as follows.
Since 
\begin{equation}
\begin{aligned}
(\tilde{g}(\btheta+\beta u)^2-\tilde{g}(\btheta))^2 &= [\tilde{g}(\btheta+\beta u)-\tilde{g}(\btheta) - \beta\langle\nabla g(\btheta),u\rangle+\beta\langle\nabla g(\btheta)\rangle]^2\\
&\leq 2(g(\btheta+\beta u)-g(\btheta)+\epsilon_{\btheta+\beta u}-\epsilon_{\btheta})^2+2\beta^2\langle \nabla g(\btheta),u\rangle^2\\
\end{aligned}
\end{equation}
and $|\epsilon_{\btheta+\beta u}-\epsilon_{\btheta}| \leq 2\epsilon$,
\begin{equation}
[\tilde{g}(\btheta+\beta u)-\tilde{g}(\btheta)]^2 \leq 2(\frac{\beta}{2}L_1(g)||u||^2)^2 + 4{\beta}L_1(g)||u||^2 \epsilon + 8\epsilon^2 + 2\beta^2\langle \nabla g(\btheta),u\rangle^2
\end{equation}
Take expectation over u, and with Theorem 3 in ~\cite{nesterov2011random}, which is $E_u(||g'(\btheta,u)\cdot u||^2) \leq (d+4)||\nabla g(x)||^2$,
\begin{equation}
\begin{aligned}
E_u(||\hat{g}(\btheta)||^2) &\leq \frac{\beta^2}{2}L_1^2(g)E_u(||u||^6)+2E_u(||g'(\btheta,u)\cdot u||^2) + 4{\beta^2}L_1(g)\epsilon E_u(||u||^4) + 8\epsilon^2 E_u(||u||^2)\\
&\leq \frac{\beta^2}{2}L_1^2(g)(d+6)^3+2(d+4)||\nabla g(\btheta)||^2 + 4\beta \epsilon L_1(g)(d+4)^2 + 8\epsilon^2 d.
\end{aligned}
\end{equation}
With $\epsilon\sim O(\delta^2\beta)$, we could bound $E_u(||\tilde{g}(\btheta)||^2)$:
\begin{equation}
E_u(||\hat{g}(\btheta)||^2) \leq \frac{\beta^2}{2}L_1^2(g)(d+6)^3+2(d+4)||\nabla g(\btheta)||^2 + 4\beta^2 L_1(g)(d+4)^2 \delta^2 + 8\beta^2 d\delta^2.
\end{equation}
We use the result that
\begin{equation}
||\nabla g(\btheta)||^2 \leq 2||\nabla g_\beta(\btheta)||^2+\frac{\beta^2}{2}L_1^2(g)(d+4)^2,
\end{equation}
which is proved in ~\cite{nesterov2011random} Lemma 4.

Therefore, since $(n+6)^3+2(n+4)^3\leq 3(n+5)^3$, we could get
\begin{equation}
\begin{aligned}
E_u(||\hat{g}(\btheta)||^2) &\leq \frac{\beta^2}{2}L_1^2(g)(d+6)^3+2(d+4)||\nabla g(x)||^2+2(d+4)||\nabla g(\btheta)||^2\\ 
&+ 4\beta^2 L_1(g)(d+4)^2 \delta^2 + 8\beta^2 d\delta^2\\
								 &\leq \frac{\beta^2}{2}L_1^2(g)(d+6)^3+2(d+4)(2||\nabla g_\beta(x)||^2+\frac{\beta^2}{2}L_1^2(g)(d+4)^2)\\
                                 &+ 4\beta^2 L_1(g)(d+4)^2 \delta^2 + 8\beta^2 d\delta^2\\
                                 &\leq 4(d+4)||\nabla g_\beta(x)||^2+\frac{3\beta^2} {2}L_1^2(g)(d+5)^3 + 4\beta^2 L_1(g)(d+4)^2 \delta^2 + 8\beta^2 d\delta^2
\end{aligned}
\end{equation}

Therefore, since $g_\beta(\btheta)$ has Lipshcitz-continuous gradient:
\begin{equation}
 |g_\beta(\btheta_{k+1}) - g_\beta(\btheta_k) + \alpha \langle \nabla g_\beta(\btheta_k), \hat{g}_\beta(\btheta_k) \rangle| \leq \frac{1}{2}\alpha^2 L_1(g_\beta) ||\hat{g}_\beta(\btheta_k)||^2
\end{equation}
so that 
\begin{equation}
g_\beta(\btheta_{k+1}) \leq g_\beta(\btheta_k) -  \alpha \langle \nabla g_\beta(\btheta_k), \hat{g}_\beta(\btheta_k) \rangle + \frac{1}{2}\alpha^2 L_1(g_\beta) ||\hat{g}_\beta(\btheta_k)||^2 .
\end{equation}
Since
\begin{equation}
\begin{aligned}
E_u(\hat{g}(\btheta_k)) &= \frac{1}{\kappa}\int_E \frac{g(\btheta + \beta u)-g(\btheta)+\epsilon_{\btheta+\beta u}-\epsilon_{\btheta}}{\beta} u e^{-\frac{1}{2}||u||^2} du\\
& = \nabla g_\beta(\btheta_k)+\frac{1}{\kappa}\int_E \frac{\epsilon_{\btheta+\beta u}-\epsilon_{\btheta}}{\beta} u e^{-\frac{1}{2}||u||^2} du\\
& \leq \nabla g_\beta(\btheta_k) + \frac{2\epsilon}{\beta} n^{1/2} \cdot \mathbb{1}
\end{aligned}
\end{equation}
where $\mathbb{1}$ is a all-one vector,
taking the expectation in $u_k$, we obtain
\begin{equation}
\begin{aligned}
E_{u_k}(g_\beta(\btheta_{k+1})) &\leq g_\beta(\btheta_k) - \alpha_k||\nabla g_\beta(\btheta_k)||^2 + \alpha_k\langle\nabla g_\beta(\btheta_k),\frac{2\epsilon}{\beta} n^{1/2} \cdot \mathbb{1} \rangle + \frac{1}{2}\alpha_k^2 L_1(g_\beta) E_{u_k}||\hat{g}_\beta(\btheta_k)||^2\\
E_{u_k}(g_\beta(\btheta_{k+1})) &\leq g_\beta(\btheta_k) - \alpha_k||\nabla g_\beta(\btheta_k)||^2+ \alpha_k\frac{2\epsilon}{\beta} n^{1/2} ||\nabla g_\beta(\btheta_k)|| \\&
+ \frac{1}{2}\alpha_k^2 L_1(g)(4(d+4)||\nabla g_\beta(\btheta_k)||^2+\frac{3\beta^2}{2}L_1^2(g)(d+5)^3+ 4\beta^2 L_1(g)(d+4)^2 \delta^2 + 8\beta^2 d\delta^2).
\end{aligned}
\end{equation}
Choosing $\alpha_k=\hat{\alpha}=\frac{1}{4(d+4)L_1(f)}$, we obtain
\begin{equation}
\begin{aligned}
E_{u_k}(g_\beta(\btheta_k+1)) &\leq g_\beta(\btheta_k) - \frac{1}{2}\hat{\alpha}||\nabla g_\beta(\btheta_k)||^2+\alpha_k\frac{2\epsilon}{\beta} d^{1/2}||\nabla g_\beta(\btheta_k)||+\frac{3\beta^2}{64}L_1(g)\frac{(d+5)^3}{(d+4)^2} \\
&+ \frac{\beta^2}{8} \delta^2+\frac{\beta^2d}{4(d+4)^2L_1(g)}\delta^2.
\end{aligned}
\end{equation}
Since $(d+5)^3\leq(d+8)(d+4)^2$, taking expectation over $\mathcal{U}_k$, where $\mathcal{U}_k =\{u_1,u_2,\dots,u_k\}$, we get
\begin{equation}
\phi_{k+1} \leq \phi_k - \frac{1}{2}\hat{\alpha}E_{\mathcal{U}_k}(||\nabla g_\beta(\btheta_k)||^2)+  \frac{3\beta^2(d+8)}{64}L_1(g) +  \frac{\beta^2}{8} \delta^2+\frac{\beta^2d}{4(d+4)^2L_1(g)}\delta^2 +\hat{\alpha}d^{1/2}E_{\mathcal{U}_k}(||\nabla g_\beta(\btheta_k)||)\delta^2,
\end{equation}
where $\phi_{k}=E_{\mathcal{U}_{k-1}(g(\btheta_k))},k\geq 1$ and $\phi_0 = g(\btheta_0)$.

Assuming $g(x)\geq g^*$, summing over k and divided by N+1, we get
\begin{equation}
\begin{aligned}
\frac{1}{N+1} \sum\limits_{k=0}^N E_{\mathcal{U}_k}(||\nabla g_\beta(\btheta_k)||^2) &\leq 8(d+4)L_1(g)[\frac{g(x_0)-g^*}{N+1}+\frac{3\beta^2(d+8)}{16}L_1(g)+\frac{\beta^2}{8} \delta^2\\
&+\frac{\beta^2d}{4(d+4)^2L_1(g)}\delta^2 + \frac{1}{N+1}\sum\limits_{k=0}^N E_{\mathcal{U}_k}(||\nabla g_\beta(\btheta_k)||) \delta^2 ] .
\end{aligned}
\end{equation}
Clearly, $\frac{1}{N+1}\sum\limits_{k=0}^N E_{\mathcal{U}_k}(||\nabla g_\beta(\btheta_k)||) \leq \delta^2$.

Since $\vartheta _k^2=E_{\mathcal{U}_k}(||\nabla g(\btheta_k)||^2)\leq 2 E_{\mathcal{U}_k}(||\nabla g_\beta(\btheta_k)||^2) + \frac{\beta^2(d+4)^2}{2}L_1^2(g)$, $\vartheta_k^2$ is in the same order oas$E_{\mathcal{U}_k}(||\nabla g_\beta(\btheta_k)||^2)$.
In order to satisfy $\frac{1}{N+1}\sum\limits_{k=0}^N\vartheta_k^2\leq \delta^2$,
we need to choose $\beta \leq O(\frac{\delta}{dL_1(g)})$, then N is bounded by $O(\frac{d}{\delta^2})$.

\end{document}